\documentclass{article}
\usepackage{spconf,amsmath,graphicx}
\usepackage{amssymb}
\usepackage{multirow}
\usepackage{booktabs}
\usepackage{setspace}
\usepackage{color, colortbl}
\usepackage[flushleft]{threeparttable}
\usepackage{tikz}
\pgfrealjobname{multiscale_denoise}
\usetikzlibrary{arrows,fadings,patterns,shapes}
\usetikzlibrary{decorations.pathmorphing}
\usetikzlibrary{decorations.pathreplacing}

\newcommand{\vv}[1]{\mathbf{#1}}
\newcommand{\mt}{\mathrm{T}}
\definecolor{Gray}{gray}{0.9}

\title{FAST, TRAINABLE, MULTISCALE DENOISING}
\name{Sungjoon Choi, John Isidoro, Pascal Getreuer, Peyman Milanfar}
\address{Google Research\\
\{sungjoonc, isidoro, getreuer, milanfar\}@google.com}
\makeatletter
\def\endthebibliography{%
  \def\@noitemerr{\@latex@warning{Empty `thebibliography' environment}}%
  \endlist
}
\makeatother
\begin{document}
\maketitle
%

\begin{abstract}
Denoising is a fundamental imaging problem.
Versatile but fast filtering has been demanded for mobile camera systems.
We present an approach to multiscale filtering which allows real-time applications on low-powered devices.
The key idea is to learn a set of kernels that upscales, filters, and blends patches of different scales guided by local structure analysis.
This approach is trainable so that learned filters are capable of treating diverse noise patterns and artifacts.
Experimental results show that the presented approach produces comparable results to state-of-the-art algorithms while processing time is orders of magnitude faster.
\end{abstract}
\section{Introduction}
\label{sec:intro}

Image denoising is known as a challenging problem that has been explored for many decades.
Patch matching methods~\cite{Buades05, elad_sparse06, dabov2007image, Kervrann08, zhang_PR2010} exploit repetitive textures and produce high quality results thanks to more accurate weight computation than bilateral filtering~\cite{Smith97, Tomasi98}.
However, higher computational complexity limits their application for low-powered devices.
Recently, deep learning based approaches~\cite{NIPS2012_4686, Zhang2017} have become popular.
While trainable networks achieve generic and flexible processing capabilities, deep layers are hard to analyze and are even more computationally expensive than patch based methods, making them harder to use in real-time applications.

Meanwhile, multiscale strategies have been widely adopted for various problems in the signal processing and computer vision communities~\cite{Anderson84pyramidmethods, Gortler96, Paris2011}.
Multiscale techniques effectively increase the footprint of filter kernels while introducing minimal overhead and allow for more efficient application of filtering than fixed-scale kernels.
Consequently, it is natural to take advantage of the multiscale approach for denoising.

Our work makes two contributions.
First, we introduce a ``shallow" learning framework that trains and filters very
fast using local structure tensor analysis on color pixels.
Because it has only a few convolution layers, the set of resulting filters is easy to visualize and analyze.
Second, we cascade the learning stage into a multi-level pipeline to effectively
filter larger areas with small kernels.
In each stage of the pipeline, we train filtering that jointly
upscales coarser level $(\ell + 1)$ and denoises and blends finer level $\ell$.

\section{RELATED WORK}
\label{sec:related_work}

The influential non-local means (NLM) filtering~\cite{buades2005nonlocal} has
received great interest since its introduction. NLM generalizes bilateral
filtering by using patch-wise photometric distance to better characterize
self-similarity, but at increased computational cost. Many techniques have been
proposed to accelerate NLM~\cite{Buades05, Liu13}. \cite{Nercessian12}~uses a
multiscale approach to perform NLM filtering at each level of a Laplacian
pyramid. The pull-push NLM~\cite{Isidoro16} method constructs up and down
pyramids where NLM weights are fused separately.

Sparsity methods open a new chapter in denoising. The now classic block-matching
and 3D filtering (BM3D)~\cite{dabov2007image} based on 3D collaborative Wiener
filtering is considered to be state-of-the-art for Gaussian noise. Nonlocally
centralized sparse representation (NCSR)~\cite{DongZSL13} introduces a sparse
model that can be solved by a conventional iterative shrinkage algorithm.

Learning-based methods have also become popular in image processing recently.
Trainable nonlinear reaction diffusion (TNRD)~\cite{ChenP17} uses multi-stage
trainable nonlinear reaction diffusion as an alternative to CNNs where the
weights and the nonlinearity is trainable. Rapid and accurate image super
resolution (RAISR)~\cite{romano2017raisr} is an efficient edge-adaptive image
upscaling method that uses structure tensor features to select a filter at each
pixel from among a set of trained filters. Best linear adaptive enhancement
(BLADE)~\cite{getreuer17blade} generalizes RAISR to a two-stage shallow
framework applicable to a diverse range of imaging tasks, including denoising.

\section{FILTER LEARNING}
\label{sec:filter_learning}

We begin with BLADE filter learning. The framework in
BLADE~\cite{getreuer17blade} is formulated for filtering an image at a
single scale. We extend BLADE to a trainable multi-level filter framework for
denoising, using noisy and noise-free images as training pairs.

\par
\begin{figure}[!t]
\begin{minipage}[b]{1.0\linewidth}
  \centering
  \centerline{\includegraphics[width=8.5cm]{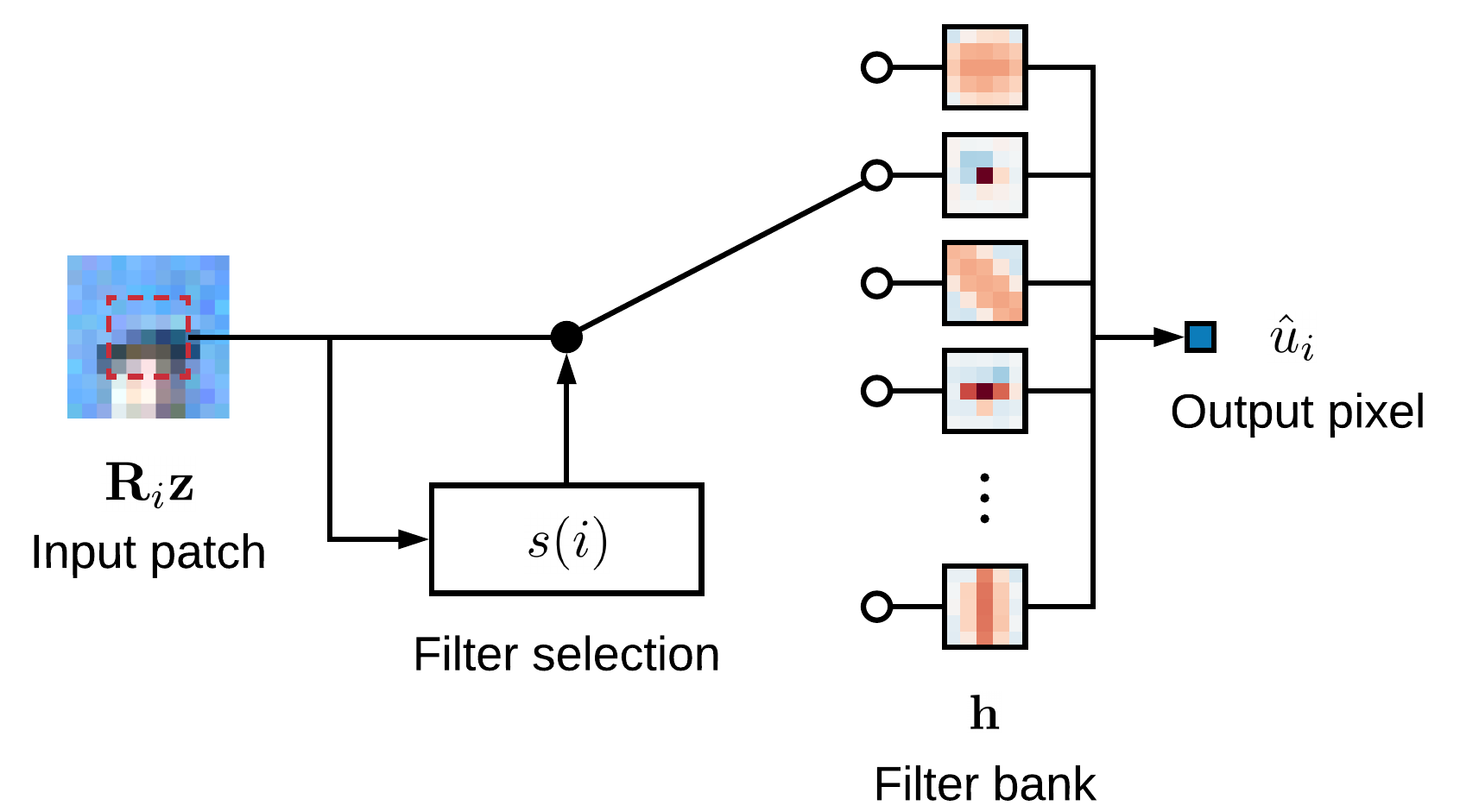}}
\end{minipage}
\caption{\label{fig:inference_diagram} Two stage spatially-adaptive filtering.
For a given output pixel $\hat{u}_i$, we only need to evaluate the one linear filter that is selected by $s(i)$.}
\end{figure}

\vspace{0.2cm}\noindent\textbf{Spatially-adaptive filtering.}
The input image is denoted by $\vv{z}$ and the value at pixel location $i \in \Omega \subset \mathbb{Z}^2$ is denoted by $z_i$.
Spatially-adaptive filtering operates with a set of linear FIR filters $\vv{h}^1, \dots, \vv{h}^K$.
$h_j^k$ denotes a filter value of $\vv{h}^k$, where $j \in F \subset \mathbb{Z}^2$ and $F$ is the footprint of the filter.
The main idea of BLADE is that a different filter is selected by a function
$s:\Omega \to \left\{1, \dots, K\right\}$ for each output pixel,
\begin{equation}\label{e:inference}
\hat{u}_i = \sum_{j \in F} h_j^{s(i)} z_{i+j}.
\end{equation}
Or in vector notation, the $i$th output pixel is
\begin{equation}
\hat{u}_i = (\vv{h}^{s(i)})^\mt \vv{R}_i \vv{z},
\end{equation}
where $\vv{R}_i$ is an operator that extracts a patch centered at $i$.
Fig.~\ref{fig:inference_diagram} depicts the two stage pipeline that adaptively selects one filter from a linear filterbank for each pixel.

\par
\begin{figure}[!t]
\begin{minipage}[b]{0.32\linewidth}
  \centering
  \centerline{\includegraphics[width=\linewidth]{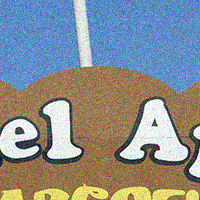}}
  \centerline{(a)}\medskip
\end{minipage}
\hfill
\begin{minipage}[b]{0.32\linewidth}
  \centering
  \centerline{\includegraphics[width=\linewidth]{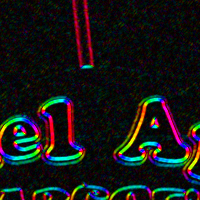}}
  \centerline{(b)}\medskip
\end{minipage}
\hfill
\begin{minipage}[b]{0.32\linewidth}
  \centering
  \centerline{\includegraphics[width=\linewidth]{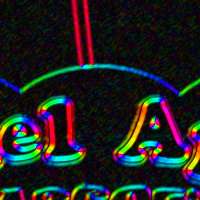}}
  \centerline{(c)}\medskip
\end{minipage}
\caption{Visualization of structure analysis where estimated orientations and
strengths are mapped to hue and value, respectively.
(a) Input image.
(b) Structure analysis of \cite{romano2017raisr}.
(c) Our structure analysis. Note that
strong edges are not detected in (b) which results in a blurry reconstruction.}
\label{fig:structure}
\end{figure}

\vspace{0.2cm}\noindent\textbf{Filter selection.}
Filter selection should segregate input patches so that the relationship to the
corresponding target pixels is well-approximated by a linear estimator, while
keeping a manageable number of filters. Filter selection should also be robust
to noise, and computationally efficient. In this light, we use features of the
structure tensor. While \cite{romano2017raisr,getreuer17blade} use the
structure tensor of the image luma channel, we find that analysis of luma alone
occasionally misses key structures that are visible in color, as shown in
Fig.~\ref{fig:structure}. We find it beneficial for denoising to compute a
structure tensor jointly using all color channels, as suggested previously for
instance by Weickert~\cite{weickert1997coherence}.
Structure tensor analysis provides a robust local
gradient estimate by principal components analysis (PCA) of the gradients over
pixel $i$'s neighborhood $N_i$, as the minimizer of
\begin{equation}
\mathop{\mathrm{arg\,min}}_{\vv{a}} \sum_{j \in N_i} w_j^i \left(\vv{a}^\mt \vv{g}_j\right)^2
\end{equation}
where $\vv{g}_j$ is the gradient at pixel location $j$ and $w_j^i$ is a
spatial weighting. With a $2 \times 3$ Jacobian matrix for color-wise gradients
\begin{equation}
\vv{G}_j =
\begin{bmatrix}
\vv{g}_j^R & \vv{g}_j^G & \vv{g}_j^B
\end{bmatrix},
\end{equation}
we now find a unit vector $\vv{a}$ minimizing
\begin{equation}
\sum_j w_j^i \left\| \vv{a}^\mt \vv{G}_j\right\|^2
= \vv{a}^\mt \left( \sum_j w_j^i \vv{G}_j \vv{G}_j^\mt \right) \vv{a}
= \vv{a}^\mt \vv{T}_i \vv{a}.
\end{equation}
The spatially-filtered structure tensor $\vv{T}_i$ is
\begin{equation}
\vv{T}_i = \sum_c \sum_j w_j^i \begin{bmatrix}
g_{x,j}^c \, g_{x,j}^c & g_{x,j}^c \, g_{y,j}^c \\
g_{x,j}^c \, g_{y,j}^c & g_{y,j}^c \, g_{y,j}^c
\end{bmatrix}\end{equation}
where $c \in \{R,G,B\}$ and $(g_{x,j}^c, g_{y,j}^c)^\mt = \vv{g}_j^c$.
For each pixel $i$,
eigenanalysis of the $2 \times 2$ matrix $\vv{T}_i$ explains the variation in the gradients along the principal directions.
The unit vector $\vv{a}$ minimizing $\vv{a}^\mt \vv{T}_i \vv{a}$ is the
eigenvector of $\vv{T}_i$ corresponding to the smallest eigenvalue, which forms
the orientation feature. The square root of the larger eigenvalue $\lambda_1$ is
a smoothed estimate of the gradient magnitude~\cite{FengOrientation}. In
addition, we use coherence $\left( \sqrt{\lambda_1} - \sqrt{\lambda_2} \right) /
\left( \sqrt{\lambda_1} + \sqrt{\lambda_2} \right)$ from the eigenvalues
$\lambda_1 \geq \lambda_2$, which ranges from 0 to 1 and characterizes the
degree of local anisotropy. We use these three features for filter selection $s$
to index a filter in the filterbank.

\begin{figure*}[!t]
\centering
\includegraphics[width=0.95\textwidth]{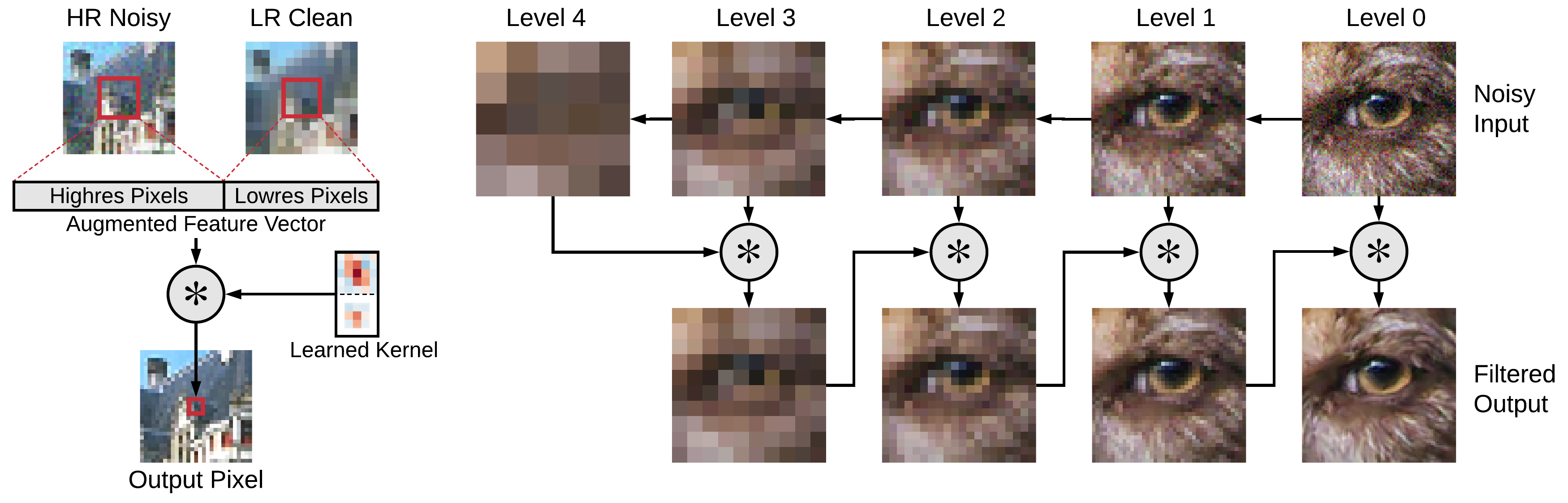}
\caption{Overview of multiscale denoising.
(left) A filter kernel is learned so that it upscales a coarser-level filtered image, filters a finer-level noisy image, and blends them into a target pixel.
(right) Combined together, cascaded learned filters form a large and irregular kernel and effectively remove noise on variable structure.}
\label{fig:multiscale}
\end{figure*}

Given a target image $\vv{u}$ and its pixel value $u_i$ at pixel location $i$, we formulate filter learning as
\begin{gather}\label{e:train1}
\mathop{\mathrm{arg\,min}}_{\vv{h}^1, \dots, \vv{h}^K}\left\|\vv{u} -
\hat{\vv{u}}\right\|^2 \\
\label{e:train2}
\begin{aligned}
\left\|\vv{u} - \hat{\vv{u}}\right\|^2
&= \sum_{k=1}^{K} \sum_{\substack{i \in \Omega : \\ s(i)=k}} \left|u_i - \hat{u}_i\right|^2 \\
&= \sum_{k=1}^{K} \sum_{\substack{i \in \Omega : \\ s(i)=k}} \left|u_i - (\vv{h}^k)^\mt \vv{R}_i \vv{z}\right|^2
\end{aligned}
\end{gather}
which amounts to a multivariate linear regression for each filter $\vv{h}^k$,
described in detail in \cite{getreuer17blade}. The above training and filtering
steps are repeated for each color channel\footnote{To denoise color images,
images are converted to YCbCr (ITU-R BT.601). We train filters separately on Y,
Cb, and Cr while using the same filter selector $s(\cdot)$ to capture
channel-specific noise statistics.}.

\section{MULTISCALE DENOISING}
\label{sec:multiscale}

In this section, we describe multiscale denoising.
The overview of the pipeline is described in Fig.~\ref{fig:multiscale}.

\par\vspace{0.2cm}\noindent\textbf{Fixed-scale filtering.}
The framework described in Section~\ref{sec:filter_learning} can be
trained from pairs of noisy and clean images to perform denoising. Based on the
noise in the training data, denoisers for different kinds of noise can be
trained. For example, \cite{getreuer17blade} shows that BLADE can perform both
AWGN denoising and JPEG compression artifact removal, interpreting JPEG
artifacts as noise. Other more complex noise models or real world noise could be
learned thanks to the generic trainable framework.

\begin{figure}[!t]
\centering
\mbox{%
\beginpgfgraphicnamed{images/single_level_filters_fig}%
\input{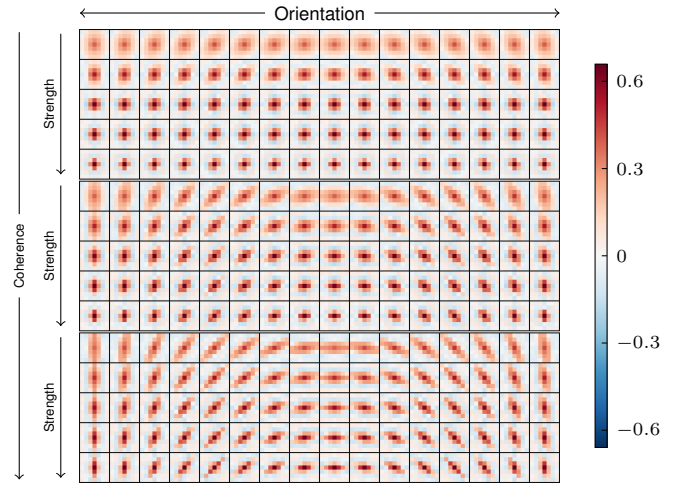}%
\endpgfgraphicnamed}
\caption{\label{fig:simple_denoiser_filters} $7\times 7$ filters for AWGN denoising with noise standard deviation 20, 16 different orientations, 5 strength values, and 3 coherence values.}
\label{fig:single_level_filters}
\end{figure}

\begin{figure}[!t]
\begin{minipage}[b]{0.49\linewidth}
  \centering
  \centerline{\includegraphics[width=\linewidth]{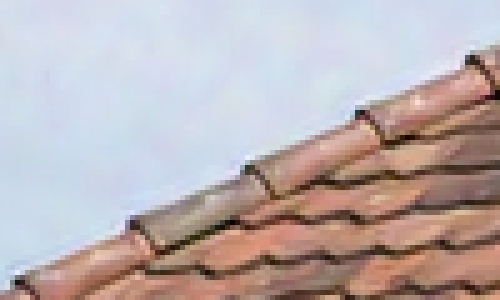}}
\end{minipage}
\hfill
\begin{minipage}[b]{0.49\linewidth}
  \centering
  \centerline{\includegraphics[width=\linewidth]{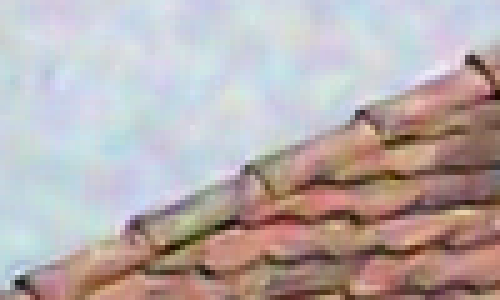}}
\end{minipage}
\caption{Results of fixed-scale denoising.
(left) Low noise input ($\sigma = 15$).
(right) High noise input ($\sigma = 50$).}
\label{fig:single_level_results}
\end{figure}

Fig.~\ref{fig:single_level_filters} visualizes learned filters for AWGN noise where $\sigma = 20$.
Fig.~\ref{fig:single_level_results} shows the denoised results with the filters trained for different noise levels.
Fixed-scale filtering is effective for the low noise level while it exhibits insufficient power for stronger noise because the footprint of the used filter ($7\times7$) is too small to compensate the noise variance.
Increasing the size of the filters is undesirable as it increases the time
complexity quadratically.

\begin{figure*}[!t]
\centering
\includegraphics[width=1.0\textwidth]{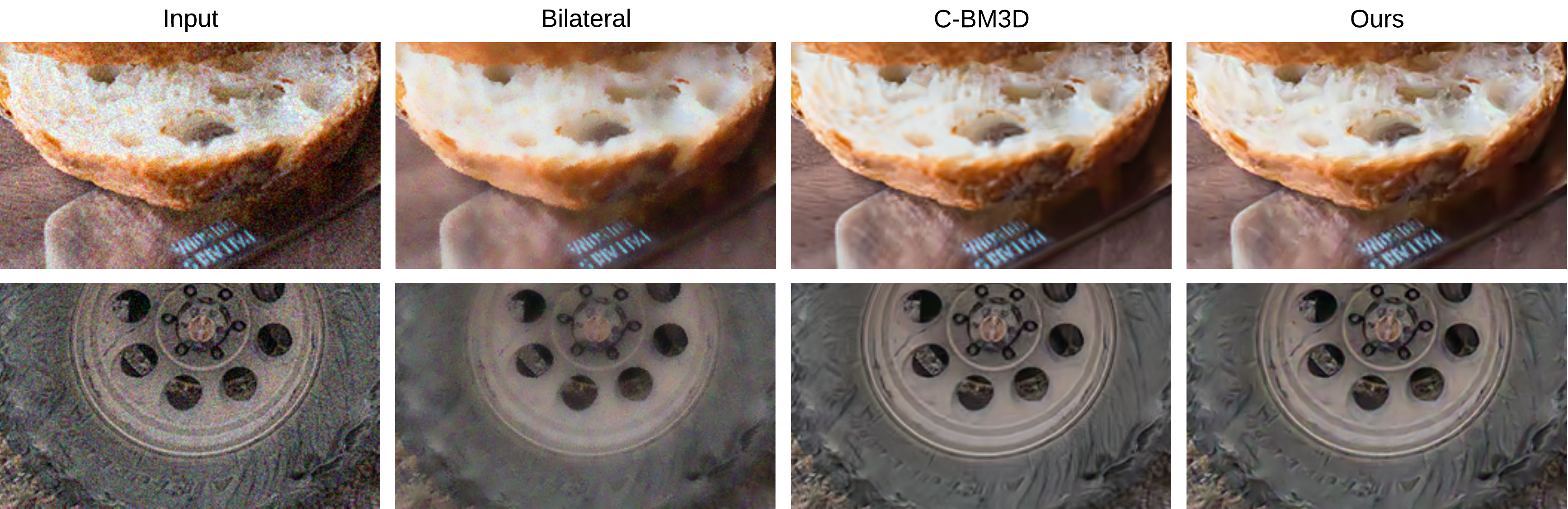}
\caption{Qualitative comparisons with noisy input $\sigma = 20$.}
\label{fig:results}
\end{figure*}

\par\vspace{0.2cm}\noindent\textbf{Multiscale filtering.}
We consider the fixed-scale denoising filter as a building block for multiscale training and inference.
We begin by taking noisy input images and forming pyramids by downsampling by factors of two.
Standard bicubic downsampling is enough to effectively reduce noise level by
half, and is extremely fast.
Pyramids of target images are constructed in the same fashion.

We start training from the second from the coarsest level $L$. To compute the output
$\vv{\hat{u}}^\ell$ at level $\ell$, the filters $\vv{f}^\ell$ upscale the next
coarser output $\vv{\hat{u}}^{\ell + 1}$ and filters $\vv{g}^\ell$ denoise and
blend with the current level's noisy input $\vv{z}^\ell$,
\begin{equation}
\hat{u}_i^\ell = \sum_{j\in F} f^{\ell, s(i)}_j \hat{u}^{\ell + 1}_{i/2+j}
+ \sum_{j\in F} g^{\ell, s(i)}_j z^\ell_{i+j},
\end{equation}
or denoting the filter pair by $\vv{h}^{\ell, k} =
\bigl[\begin{smallmatrix}
\vv{f}^{\ell, k} \\ \vv{g}^{\ell, k}
\end{smallmatrix}\bigr]$, as
\begin{equation}
\hat{u}_i^\ell = (\vv{h}^{\ell,s(i)})^\mt
\begin{bmatrix}
\vv{R}_{i/2} \vv{\hat{u}}^{\ell + 1} \\
\vv{R}_i \vv{z}^\ell
\end{bmatrix}
\end{equation}
where the base case $\vv{\hat{u}}^{L} = \vv{z}^{L}$
as illustrated in Fig.~\ref{fig:multiscale}~(a). We train filters by using input
patches from the current level $\ell$ and corresponding patches from the
filtered output at the next coarser level $\ell + 1$. Once level $\ell$ is
trained, filtered images $\vv{\hat{u}}^\ell$ are computed and then consumed
for training $\vv{h}^{\ell - 1}$ at the next finer level.

%

Overall, our shallow inference can be performed with high efficiency. Both
the selection $s$ and filtering are vectorization and parallelization-friendly
because most operations are additions and multiplications on
sequential data with few dependencies. On the Pixel~2017 phone, inference time
is 18~MP/s on CPU and 188~MP/s on GPU.

%

\section{EXPERIMENTAL RESULTS}

We have evaluated the presented pipeline on 68 test images of the Berkeley dataset~\cite{MartinFTM01}.
We used high quality images separately collected from the Internet to train a filterbank where about $2.23 \times 10^9$ pixels were consumed.
Noisy images were synthesized with an AWGN model, and then quantized and clamped at 8-bit resolution.
To get more samples, we included spatial axial flips and $90^{\circ}$ rotations of each observation and target image pair in the training set so that the filters learn symmetries.
At every level, we used 16 orientation buckets, 16 strength buckets, and 16 coherence buckets for structure analysis.
For low noise level $\sigma < 10$, the filter size of $5 \times 5$ was used for finer level and $3 \times 3$ for coarser level.
Otherwise, the filter size of $7 \times 7$ was used for finer level and $5 \times 5$ for coarser level.
The level of pyramid is set so that the noise standard deviation of the coarsest level is less than 2.

Table 1 reports the PSNR and timing of various state-of-the-art techniques.
We provided each method with the same noise variance parameter used to synthesize noisy input.
Running times were measured on a workstation with an Intel Xeon E5-1650 3.5 GHz CPU.
The results of the proposed pipeline are comparable to the state-of-the-art algorithms as shown in Fig.~\ref{fig:results} while it is orders of magnitude faster.
Per-image processing time of ours was linear to the number of pixels in the image.

\begin{table}
\begin{threeparttable}
\centering
\caption{\label{table:performance_vs_device}
Quantitative evaluation with the Berkeley dataset.}
\begin{small}
\begin{tabular}{lcccc}
\hline
\multirow{2}{*}{Method} & \multicolumn{3}{c}{PSNR (dB)} & \multirow{2}{*}{Time (s)} \\
& $\sigma=15$ & $\sigma=25$ & $\sigma=50$ &  \\
\hline
BM3D~\cite{dabov2007image} & 30.87 & 28.20 & 24.63 & 47.2 \\
K-SVD~\cite{elad_sparse06} & 30.66 & 27.82 & 23.80 & 35.8 \\
$\textrm{TNRD}_{7\times7}$~\cite{ChenP17} & 31.18 & 28.48 & 24.75 & 16.5 \\
\rowcolor{Gray}
C-BM3D~\cite{DabovFKE07} & 33.24 & 30.18 & 25.85 & 21.9 \\
\rowcolor{Gray}
Ours & 32.46 & 29.58 & 25.92 & 0.038 \\
\hline
\end{tabular}
\begin{tablenotes}
\item For methods shaded with gray, color channels were jointly denoised; otherwise the filters were independently applied on each channel.
Running times were measured on 1 MP images.
\end{tablenotes}
\end{small}
\end{threeparttable}
\end{table}

\section{CONCLUSION}
\label{sec:conclusion}

We have presented a trainable multiscale approach for denoising.
The key idea is to learn filters that jointly upscale, blend, and denoise between
successive levels.
The learning process is capable of treating diverse noise patterns and
artifacts. Experiments demonstrate that the presented approach produces results
comparable to state-of-the-art algorithms with processing time that is orders of
magnitude faster.

The presented pipeline is not perfect.
For inference, we assumed the noise level of input image is known and used the filters trained with the data of the same noise level.
There are many ways to estimate the level of noises, which can guide us to select the right filter set.
Also we assumed the noise level is uniform across pixel locations.
We believe we can characterize and model the noise response of a camera system,
and then integrate this information into filter selection.



\bibliographystyle{IEEEbib}
\bibliography{refs}

\end{document}